\documentclass[twocolumn]{wlscirep}

\usepackage[utf8]{inputenc}
\usepackage[T1]{fontenc}
\usepackage{indentfirst}
\usepackage{ragged2e}
\usepackage[font=small,labelfont=bf, justification=justified, format=plain]{caption}
\usepackage{xspace}

\usepackage{times}
\usepackage{graphicx}
\usepackage[normalem]{ulem}
\usepackage[switch]{lineno}

\usepackage{color}
\usepackage{hyperref}
\hypersetup{
	colorlinks,
	linkcolor={red!100!black},
	citecolor={blue!100!black},
	urlcolor={blue!80!black}
}
\usepackage{amssymb}
\usepackage{amsmath, bm}
\usepackage{mathrsfs}  
\usepackage{microtype}
\usepackage[font=footnotesize,skip=2pt]{caption}
\usepackage{tikz}
\newcommand*\circled[1]{\tikz[baseline=(char.base)]{
            \node[shape=circle,draw,inner sep=0.5pt] (char) {#1};}}
\newcommand{\link}{\mathcal{L}k}
\newcommand{\writhe}{\mathcal{W}r}
\newcommand{\twist}{\mathcal{T}w}

\usepackage{pict2e}

\DeclareRobustCommand{\wavesymbol}{%
  \begingroup\setlength{\unitlength}{\fontcharht\font`A}%
  \begin{picture}(1.6,1)
  \roundcap
  \thicklines
  \put(0.1, 0.8){\line(1,0){.6}}
  \put(0.7, 0.8){\line(0,-1){0.8}}
  \put(0.7, 0.0){\line(1,0){.6}}
  \end{picture}%
  \endgroup
}

\DeclareRobustCommand{\pulsesymbol}{%
  \begingroup\setlength{\unitlength}{\fontcharht\font`A}%
  \begin{picture}(1.6,1)
  \roundcap
  \thicklines
  \put(0.1, 0.0){\line(1,0){.4}}
  \put(0.5, 0.0){\line(0,1){.8}}
  \put(0.5, 0.8){\line(1,0){.6}}
  \put(1.1, 0.8){\line(0,-1){.8}}
  \put(1.1, 0.0){\line(1,0){.4}}
  \end{picture}%
  \endgroup
}

\usepackage{babel}
\usepackage{csquotes}
\usepackage{jabbrv}
\usepackage[style=nature, autocite=superscript, backend=biber]{biblatex}
  
\AtBeginBibliography{\footnotesize}
\let\cite\autocite
\defbibfilter{appendixOnlyFilter}{
  segment=1 
  and not segment=0 
}
\newcommand{\add}[1]{\textcolor{blue}{#1}}

\title{
Topology, dynamics, and control of an octopus-analog muscular hydrostat 
} 

\makeatletter
\renewcommand\AB@affilsepx{, \protect\Affilfont}
\makeatother

\author[1]{Arman Tekinalp}
\author[2]{Noel Naughton}
\author[1]{Seung-Hyun Kim}
\author[3]{Udit Halder}
\author[4]{Rhanor Gillette}
\author[1,3]{Prashant G. Mehta}
\author[5]{William Kier}
\author[1*]{Mattia Gazzola}

\affil[1]{\footnotesize{Mechanical Science and Engineering, University of Illinois at Urbana-Champaign, Urbana, Illinois}}
\affil[2]{\footnotesize{Beckman Institute for Advanced Science and Technology, University of Illinois at Urbana-Champaign, Urbana, Illinois}}
\affil[3]{\footnotesize{Coordinated Science Laboratory, University of Illinois at Urbana-Champaign, Urbana, Illinois}}
\affil[4]{\footnotesize{Department of Molecular and Integrative Physiology, University of Illinois at Urbana-Champaign, Urbana, Illinois}}
\affil[5]{\footnotesize{Department of Biology, University of North Carolina, Chapel Hill, North Carolina}}
\affil[*]{\footnotesize{mgazzola@illinois.edu}}

\date{}

\begin{abstract}
\vspace{-45pt}

Muscular hydrostats, such as octopus arms or elephant trunks, lack bones entirely, endowing them with exceptional dexterity and reconfigurability. Key to their unmatched ability to control nearly infinite degrees of freedom is the architecture into which muscle fibers are weaved. Their arrangement is, effectively, the instantiation of a sophisticated mechanical program that mediates, and likely facilitates, the control and realization of complex, dynamic morphological reconfigurations. Here, by combining medical imaging, biomechanical data, live behavioral experiments and numerical simulations, we synthesize a model octopus arm entailing $\sim$200 continuous muscles groups, and begin to unravel its complexity. We show how 3D arm motions can be understood in terms of storage, transport, and conversion of topological quantities, effected by simple muscle activation templates. These, in turn, can be composed into higher-level control strategies that, compounded by the arm’s compliance, are demonstrated in a range of object manipulation tasks rendered additionally challenging by the need to appropriately align suckers, to sense and grasp. Overall, our work exposes broad design and algorithmic principles pertinent to muscular hydrostats, robotics, and dynamics, while significantly advancing our ability to model muscular structures from medical imaging, with potential implications for human health and care.

\end{abstract}

\begin{document}
\maketitle

By forgoing hard skeletal support in favor of three-dimensional, densely-packed fiber structures, muscular hydrostats bypass rigid kinematic constraints to achieve unparalleled dexterity and reconfigurability \cite{Kier:1985}. It is thus perhaps not surprising that these solutions have independently evolved across taxa and environments, from the prehensile tongues of giraffes \cite{Pretorius:2016} to the nimble trunks of elephants \cite{Shoshani:1998} or arms of octopuses \cite{Kier:1985}, long fascinating biologists, mathematicians, and engineers alike. Key to these organs’ dexterity are the architectural motifs into which muscle fibers are weaved and connected together. Indeed, these muscular arrangements embody sophisticated mechanical programs that translate 1D muscle contractions into complex 3D deformations, possibly relieving the neural system of taxing computations\cite{Pfeifer:2006}. Despite much interest and broad technological implications, distilling design and control principles from these intricate systems has proved challenging.

Kinematic studies of animal tentacles \cite{Kier:1997}, arms \cite{Kennedy:2020, Richter:2015, Hanassy:2015, Zelman:2013}, tongues \cite{Smith:1984, Wainwright-1:1992}, and trunks \cite{Dagenais:2021} have provided useful characterizations, but cannot fully elucidate the relationship between control, actuation, and dynamics. Initial insight into these relations has been provided by a series of pioneering electromyography recordings in octopus arms 
\cite{Gutfreund:1998, Sumbre:2001,Sumbre:2005,Sumbre:2006}. These studies illustrate how simple templates of electrical activity, consisting of traveling and colliding waves, underlie bend propagation \cite{Gutfreund:1998, Sumbre:2001} and joint formation during planar fetching motions \cite{Sumbre:2005,Sumbre:2006}. Nonetheless, full spatial and temporal activation patterns at the individual muscle level, particularly important for decoding the organization of 3D movements, remain inaccessible. 
Robophysical approaches, despite tremendous progress 
\cite{bao2020trunk,
laschi2012soft, 
walker2005continuum, 
kang2016embodiment,  
martinez2013robotic, 
wehner2016integrated}, have also struggled to make inroads, stymied by the lack of advanced materials able to replicate the architecture and performance of natural muscular hydrostats. Within this context, \textit{in-silico} approaches can complement \textit{in-vivo} and robotic ones, allowing us to computationally explore the functioning of these systems. 
However, in spite of advances in the modeling of soft and heterogeneous structures 
\cite{VanLeeuwen:1997, Chiel:1992, liang2006finite, Yekutieli:2005-1, Kaczmarski:2023, Chang:2023}, no description has yet reached the maturity necessary to capture the extraordinary complexity of muscular hydrostats.

\begin{figure*}[th!]
    \centering
    \includegraphics[width=\textwidth]{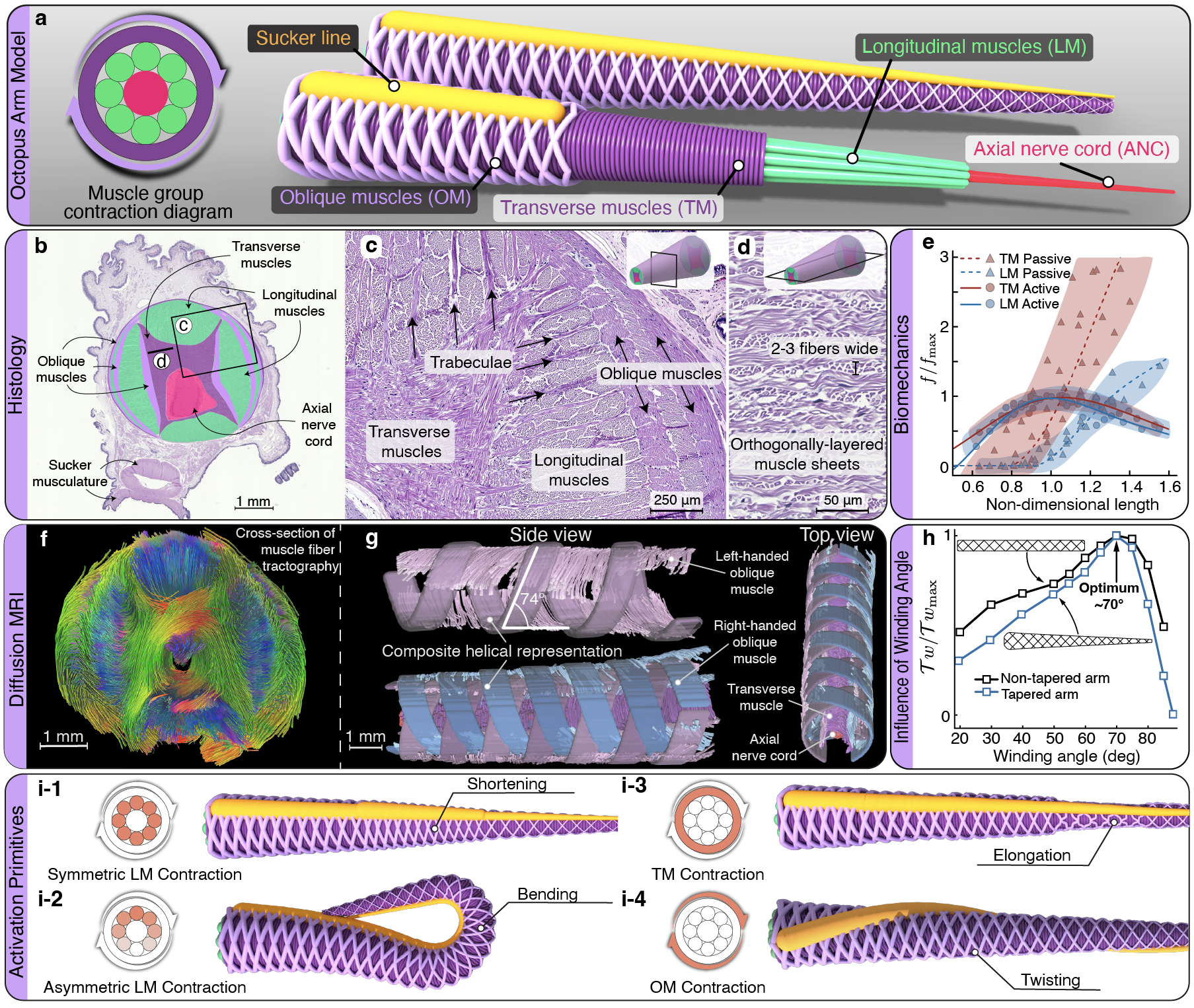}%
    \caption{
    \textbf{Computational modeling of an octopus arm from histological, biomechanical, and diffusion MRI tractography data.}
    \textbf{(a)} Cosserat rods are assembled together reflecting the octopus arm architecture.
    \textbf{(b)} Transverse slice of \textit{Octopus rubescens} arm (H\&E stain)
    with key muscle groups and anatomical features labeled. Histology insets show: 
    \textbf{(c)} extension through longitudinal muscles by trabeculae. These act as force transmission pathways to the arm’s outer layers, allowing transverse muscles to radially squeeze the arm; 
    \textbf{(d)} frontal slice of transverse muscles showing alternating arrays of orthogonal fibers, enabling transversely isotropic and axially decoupled stress generation. 
    \textbf{(e)} Force-length relationships of longitudinal and transverse muscles (solid lines) fitted to experimental data of an \textit{Octopus vulgaris} reported by Zullo et al.\cite{zullo2022octopus}. 
    \textbf{(f)} Diffusion MRI tractography of \textit{Octopus rubescens} arm showing 3D muscle-fiber arrangement. 
    \textbf{(g)} Segmentation of muscle groups allows identification of key morphological features for inclusion in our model such as the winding angle of the oblique muscle layers.
    \textbf{(h)} Effect of oblique muscle winding angle on twisting performance for both non-tapered and tapered arms. Simulations show that maximum twist is generated at a winding angle of $70^\circ$.  
    \textbf{(i)} Motion primitives arise from individual muscle groups contractions: shortening (symmetric LM), bending (asymmetric LM), elongation (TM), and twisting (OM). 
    }
    \label{fig:octopus_primitives}
    \vspace{-14pt}
\end{figure*}

Here, based on histological assays, diffusion MRI tractography, biomechanical data and live behavioral experiments, we numerically instantiate an octopus arm out of soft, active, three-dimensionally intertwined model fibers, and begin to unravel its complexity.
This model allows the direct activation of different muscle groups (individually or in concert) to effect naturally observed 3D motions\cite{Kennedy:2020}, and is used here to elucidate principles of control, revealing surprising simplicity. 
Basic templates of muscle (co-)activations are related to the storage, conversion, transport, and release of three topological/geometric descriptors which, mediated by the arm’s compliance, dynamically unfold into complex morphological reconfigurations. 
The intuitive composition of such templates allows orchestrating high-level tasks, whereby, for example, an arm can squeeze through a crevice with its suckers exposed outwards for sensing, reach for an object on the other side, realign the suckers inwards to grasp it, manipulate it, and retrieve it. 
Further compounding the robustness of this approach, the interplay between the compliant arm and solid interfaces is found to rectify imprecise control, allowing the arm to passively accommodate obstacles and objects of different geometries without changes in muscle actuation.

While these insights stem from a model of octopus, this work goes beyond the specific animal instance. Indeed, beside advancing our simulation abilities and understanding of muscular hydrostats, it identifies broad design and control principles using the perspectives of topology and geometry, with implications in biology, dynamics, and robotics.

\noindent\textbf{Modeling of an octopus arm.} 
We model the internal organization of the octopus arm through assemblies of Cosserat rods 
(Fig.~\ref{fig:octopus_primitives}a, Methods). These are one-dimensional elastic fibers that can undergo at every (circular) cross-section all six modes of deformation (bending, twisting, shearing, and stretching), and thus continuously deform in 3D space \cite{Gazzola:2018,Antman:2005}. 
Cosserat rods provide a convenient mathematical representation in the present context: they naturally map onto muscle fibers and groups, can actively contract at any location according to prescribed force-length relations, and can be connected together (via appropriate boundary conditions) into complex structures \cite{Zhang:2019}. 
The result is an accurate framework that intrinsically captures the heterogeneous and anisotropic nature of muscular hydrostats, as well as the distributed generation and transmission of internal loads.

To instantiate an  octopus arm \textit{in silico}, experimentally determined geometric and biomechanical properties are incorporated within our Cosserat representation. 
We begin by considering histological cross-sections of \textit{Octopus rubescens} to highlight the arm's main structural elements (Fig.~\ref{fig:octopus_primitives}a,b): a mechanically passive axial nerve cord (ANC) surrounded by longitudinal (LM), transverse (TM), and oblique muscle (OM) groups \cite{Kier:2007}.

The nerve cord (ANC) and longitudinal muscles (LM) span the full length of the arm and run parallel to it, with the nerve cord centered along the midline, and the longitudinal muscles located off-axis (Fig.~\ref{fig:octopus_primitives}a,b). 
When all longitudinal muscles contract, the arm shortens on account of the resulting axially compressive forces (Fig.~\ref{fig:octopus_primitives}i-1). 
If instead longitudinal muscles are selectively activated on one side of the arm, contractile forces result in distributed couples due to their off-axis alignment, producing bending (Fig.~\ref{fig:octopus_primitives}i-2). To morphologically and functionally recapitulate this structure, the nerve cord (ANC) is modeled as a single passive central rod (Fig.~\ref{fig:octopus_primitives}a, pink), while longitudinal muscles (LM) are represented by eight surrounding active rods (Fig. \ref{fig:octopus_primitives}a, green). 
This number (eight) stems from using rods of circular cross-section with diameters determined to approximate natural proportions (ANC occupies $\sim$10\% of the arm cross-section, $\sim$50\% for LM, see SI).
Biomechanically, the contractile and elastic behavior of longitudinal muscles, i.e. their characteristic active/passive stress-strain relationships (Fig.~\ref{fig:octopus_primitives}e), are based on muscle-specific measurements of \textit{O. vulgaris} by Zullo et al\cite{zullo2022octopus}.
Incorporating muscle specificity is critical to capture the octopus reconfigurability, where longitudinal muscles operate over a much wider length-range  
than transverse muscles \cite{zullo2022octopus, di2021beyond, Thompson:2023}.
Finally, the passive response of the nerve cord is modeled as for LM.

Transverse muscles (TM) are anatomically located between the nerve cord and longitudinal muscles (Fig.~\ref{fig:octopus_primitives}a,b). 
Their activation radially compresses the arm, causing it to extend axially due to the tissue's near-incompressibility (Fig.~\ref{fig:octopus_primitives}i-3). The cross-sectional enlargement of Fig.~\ref{fig:octopus_primitives}b reveals the intricate microstructure that enables this function. 
Emanating from the transverse muscles, thin muscular strips (trabeculae) infiltrate through the longitudinal muscles, reaching into the oblique muscles and outer connective tissue (Fig.~\ref{fig:octopus_primitives}c), forming a dense fan of tethers that distributes radial compressions over the cross-section. 
Further, transverse muscles are organized in independent sheets (two to three fibers thick), each orthogonal to the arm’s axis and alternating perpendicular fiber-alignments (frontal slice of Fig.~\ref{fig:octopus_primitives}d). 
This orthogonal packing allows individual sheets to approximately slip past each other during contractions, generating both finely localized and transversely isotopic compression forces without entanglement. 
We functionally recapitulate this axially decoupled, interdigitated microstructure by modeling the transverse muscles as a concatenated series of contractile rings (Fig.~\ref{fig:octopus_primitives}a, dark purple) enveloping the longitudinal muscles (green). 
Concurrently, we enhance the Cosserat formalism to capture the effect of intramuscular pressure. This is generated by transverse contractions that squeeze and elongate the longitudinal elements LM and ANC. 
The pressure model is detailed in the Methods, with quantitative validations against squid tentacle strike experiments \cite{VanLeeuwen:1997} provided in the SI. 
Ring dimensions are again based on the relative cross-sectional area occupied by transverse muscles ($\sim$20\%), while their contractile and elastic properties reflect the characterization of Zullo et al.\cite{zullo2022octopus} (Fig.~\ref{fig:octopus_primitives}e). 

Oblique muscles (OM) are helically arranged around the arm and, upon contraction, twist it (Fig.~\ref{fig:octopus_primitives}i-4). To visualize this non-planar architecture, we performed high-resolution, diffusion-weighted Magnetic Resonance Imaging (dMRI) of an \textit{O. rubescens} arm using a high-field 9.4 Tesla scanner (Methods). dMRI measures the direction-dependent diffusion of water in material samples \cite{Mori:2002}. In fibrous tissue, water molecules preferentially diffuse along fibers, thus encoding structural information into the dMRI signal. 
Tractography can then be employed to directly extract fibers’ spatial organization, synthesizing the three-dimensional architecture of the tissue\cite{Gaige:2007}.
Applied here for the first time to the octopus, dMRI tractography reveals the volumetric, muscular organization of the arm (Fig.~\ref{fig:octopus_primitives}f,g), expanding upon the architectural motifs gleaned from local, two-dimensional histology, and further guiding our modeling intuition (all imaging data are openly distributed, see Data Access statement). 
The oblique musculature is organized into three layers, external, medial, and internal, on both sides of the arm. Consistent with previously reported microscopy\cite{Kier:2007}, the handedness of fibers alternates by layer and is opposite to its contralateral pair (i.e. the external and internal layers on one side have the same handedness as the medial layer on the opposite side). Similarly-handed layers form a composite helical system (Fig.~\ref{fig:octopus_primitives}g), whereby forces produced by contractions on one side of the arm are transmitted, through the connective tissue found at the top (aboral) and bottom (oral) of the arm to the opposite side, enabling twisting motions. 
We approximate this infrastructure via two sets of four opposite-handed helical strands (R-OM and L-OM, Fig.~\ref{fig:octopus_primitives}a, bright purple) wrapping along the full length of the arm at a 74$^\circ$ angle from the base to match tractography measurements. 
Notably, we find in simulations (Fig.~\ref{fig:octopus_primitives}h) that this winding angle ($\sim70^\circ$) maximizes twist generation upon oblique muscle contraction, providing a potential mechanistic rationale for the evolution of this particular arrangement, as well as a useful design guideline for engineers (see SI for discussion).
The employed number of rods (eight) approximates the cross-sectional area proportions ($\sim$20\% for OM). Biomechanical force-length relationships for oblique muscles do not exist, and we use the LM data of Fig.~\ref{fig:octopus_primitives}e instead.

\begin{figure*}[t!]
    \centering
    \includegraphics[width=\textwidth]{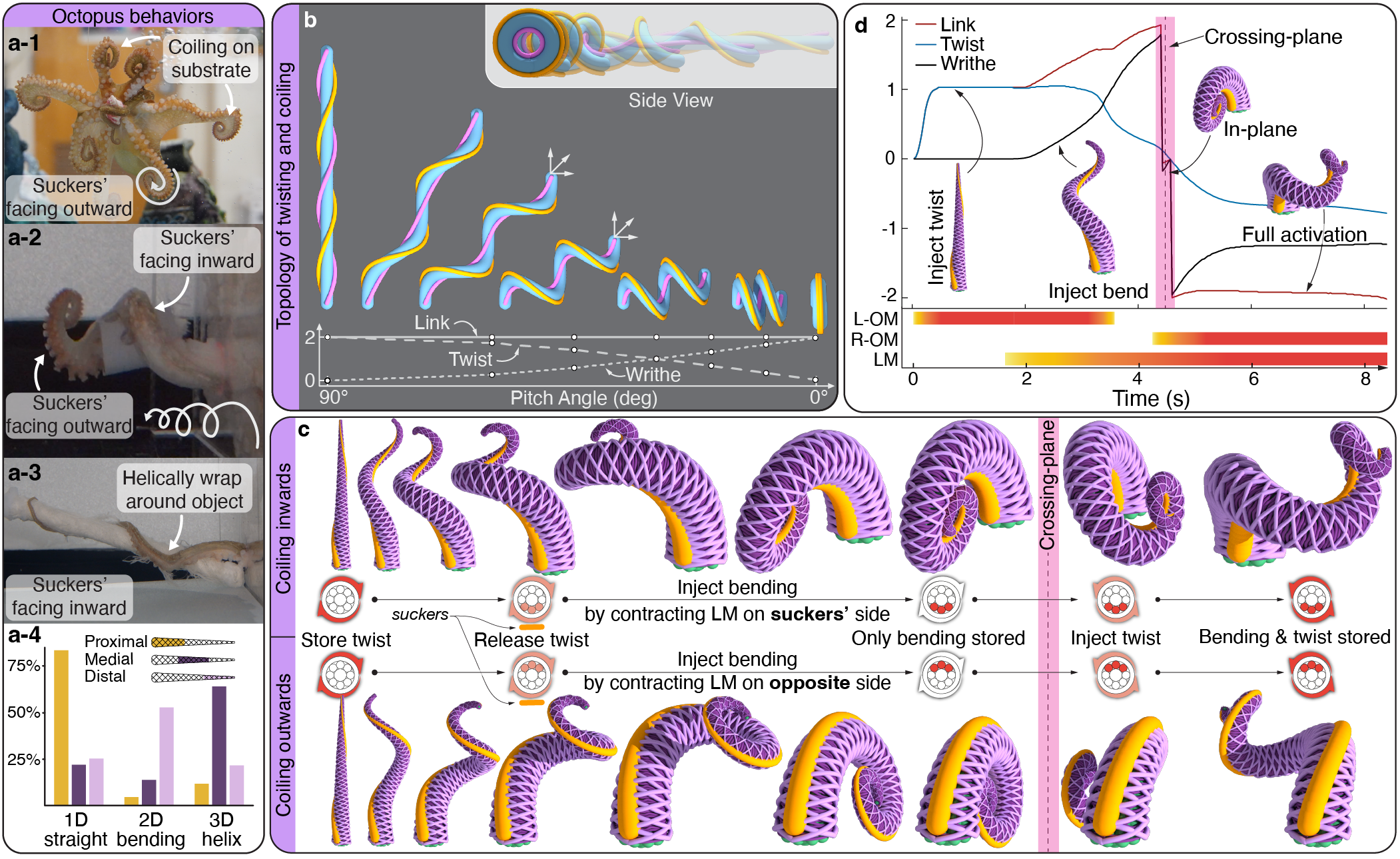}
    \caption{
    \textbf{Topological and dynamic interpretation of an octopus arm.}
    \textbf{(a)} Example images of an octopus  coiling its arms against a substrate (a-1) and forming 3D helical structures (a-2, a-3)  during object grasping and manipulation. (a-4) Video analysis in controlled conditions \add{(SI Video 1)} reveals the distribution of an octopus arms deformation modes during object grasping and manipulation.
    \textbf{(b)}
    Illustration of CFW Theorem at work: single rod (blue) and associated auxiliary curves (yellow/pink) are twisted twice ($\twist=2$) injecting constant link ($\link=2$).
    As pitch angle decreases ($90^\circ \rightarrow 0^\circ$), twist is converted into writhe, reconfiguring the rod’s morphology from straight $(\twist = 2, \writhe=0)$ to helical to planar spiral $(\twist \rightarrow 0, \writhe \rightarrow 2)$.
    \textbf{(c)} 
    Dynamic deformation of a helically coiling arm. A straight arm stores twist by contracting its L-OM, forms helices by additionally contracting LM, before folding into a spiral after relaxing L-OM. Re-injecting twist (R-OM), while keeping LM contracted, extends the arm telescopically and perpendicularly. (Top row) Contraction of LM adjacent to the suckers aligns them inwards. (Bottom row) LM activation on the opposite side aligns suckers outwards \add{(SI Video 2)}.
    \textbf{(d)} Evolution of $\link$, $\writhe$, $\twist$  during the reconfiguration of (c, top row). Pink shading denotes a transition plane where the handedness of the arm changes as it crosses through itself twice (two loops), causing $\writhe$ and $\link$ to decrease by four \cite{Fuller:1978}.
    }
    \label{fig:combined_primitives}
    \vspace{-14pt}
\end{figure*}

Fully assembled, our model arm consists of 197 rods (1 ANC, 8 LM, 4 R-OM + 4 L-OM, and 180 TM rods) and, in keeping with measurements of \textit{O. rubescens}\cite{Chang:2020}, has a length of 20 cm, a diameter at the base of 24 mm diameter, and a tapering angle of 87$^{\circ}$.
Rods are bound together via distributed boundary conditions that approximate the passive elastic properties of the extracellular matrix (details in SI). 
Each muscular rod can locally contract at any position along its length, generating axially compressive forces that are mechanically translated by the arm's architecture into three-dimensional dynamic motions. From this model, basic motor primitives such as bending (asymmetric LM contraction), twisting (OM contraction), shortening (symmetric LM contraction), and elongation (TM contraction) naturally arise (Fig.~\ref{fig:octopus_primitives}i). 
Observed octopus kinematics can then be connected to the actuation primitives that beget them, setting the stage for exploring their composition into complex behaviors. 

\noindent\textbf{A topological view informed by live octopus experiments.}  
Analogous to the complexity of real octopus arms, our model --- with nearly 200 rods bundled together, each able to continuously contract and elastically deform --- is highly non-linear and characterized by a vast number of degrees of freedom. 
Controlling such a system is a daunting task.

To simplify the problem and gain broader perspective unobscured by the arm’s specific details, we adopt a topological and geometric approach. 
We start by considering the arm as a single rod, the blue curve of Fig.~\ref{fig:combined_primitives}b with edges highlighted in yellow and pink, to understand its reconfiguration through the descriptors link $(\link)$, writhe $(\writhe)$, and twist $(\twist)$, a representation long employed in biology to characterize the supercoiling morphology of DNA \cite{Bauer:1980,Fuller:1978}. Link $(\link)$, a topological invariant, is the oriented crossing number (or Gauss linking integral) of the rod's centerline and one of its edges (pink auxiliary curve), averaged over all projections. 
Practically, link quantifies how much the two curves wind around each other. 
Writhe $(\writhe)$ is a global geometric quantity equivalent to the link of the centerline with itself. 
Essentially, it captures how much the rod bends and coils but does not account for the orientation of the edges. 
Twist $(\twist)$, also a geometric quantity, accounts for this orientation, measuring the total rotation of the auxiliary curve about the centerline’s tangent. 
Critically, these quantities are related through the Calugareanu-Fuller-White (CFW) theorem\cite{Fuller:1978, Calugareanu:1959}
\begin{equation}
    \link=\writhe+\twist,
    \label{eq:CFW}
\end{equation}
allowing us to understand arm reconfigurations in terms of injection, storage, and interconversion of only three quantities. This is illustrated in Fig.~\ref{fig:combined_primitives}b, where a set of helices of constant arc-length, but varying pitch angle (0 to 90 degrees), are generated under the constraint of orientations (tangents, normal, binormals) matching at their ends. 
This constraint renders the helical rod equivalent to a closed curve, in that link is conserved\cite{van:2007} (see SI). 
Initially, the straight rod $(\writhe=0)$ is twisted two full rotations $(\twist=2)$, injecting and permanently storing $\link=2$.
As the helical pitch increases, 3D coils begin to manifest. Coils are associated with writhe $(\writhe \uparrow)$, forcing the rod to untwist $(\twist \downarrow)$ since link cannot vary. 
Thus, as twist is converted into writhe, the helix axially contracts, until no twist is left $(\twist \to 0)$, only writhe remains $(\writhe \to 2)$, and the rod approaches a planar coil.

While illustrative of the CFW theorem, this reconfiguration sequence is also a minimal abstraction of a broad class of octopus behaviors. Indeed, planar coiling is commonly encountered in arms at rest or attached to a substrate (Fig.~\ref{fig:combined_primitives}a-1). To and from these configurations, arms often fold and unfold into helices over a section or full extent of the arm, particularly when engaging with objects. We quantify this via controlled behavioral experiments involving \textit{O. rubescens} interacting with different objects (Fig.~\ref{fig:combined_primitives}a, \add{SI Video 1}). 
Video analysis (Fig.~\ref{fig:combined_primitives}a-4, SI) reveals how proximal and distal sections primarily employ 1D (straight arm) and 2D (planar bend/coil) configurations, respectively, with suckers typically exposed outward (Fig.~\ref{fig:combined_primitives}a-1,2). 
However, the medial section, where the majority of object engagement occurs\cite{Grasso:2008}, systematically employs helices to orient the suckers inward towards the substrate, wrap around it, and manipulate it (Fig.~\ref{fig:combined_primitives}a-2,3). 

\noindent\textbf{Topological control via uniform muscle actuation.}
To understand actuation and control strategies underlying 3D manipulations, we connect muscle dynamics with the above compact, but static, topological view. We first note that $\link$, $\writhe$, and $\twist$ are all global quantities, suggesting that global (i.e uniform) muscle contractions over the full arm (or a section) may suffice to realize simple 3D behaviors.
Second, since $\link$, $\writhe$, and $\twist$ are related through Eq. \ref{eq:CFW}, only two of them need to be controlled, 
with $\writhe$ and $\twist$ intuitive candidates because, unlike $\link$, they are geometric descriptors that can be associated with the activity of individual muscle groups. 
Specifically, oblique muscles mediate $\twist$, while longitudinal muscles can be associated with the manipulation of $\writhe$ through bending. 

The utility and simplicity of this perspective is demonstrated in Fig.~\ref{fig:combined_primitives}c (top/left) and \add{SI Video 2}, where
we consider an initially straight, relaxed octopus arm ($\link = \writhe = \twist = 0$) that coils into a helix before folding into a spiral. 
To dynamically realize this motion, we first inject twist $(\twist = 1)$ by uniformly contracting the left-handed oblique muscles, thus storing link ($\link=1$) as a degenerate loop collapsed along the arm's midline.
Then, we unfold this loop into a helix by introducing writhe via the contraction of longitudinal muscles on one side of the arm.
Unlike the rod of Fig.~\ref{fig:combined_primitives}b, here the tip of the arm is free, allowing link to increase in response to longitudinal contractions.
This manifests in the gradual appearance of a second coil as link approaches $\link=2$. 
Based on Fig.~\ref{fig:combined_primitives}b, folding the newly formed helix into a spiral requires converting stored twist into writhe. 
To do so, we relax the oblique muscles (while keeping LM contracted), 
releasing twist into writhe (note faster growth rate in Fig.~\ref{fig:combined_primitives}d), until the spiral $(\writhe=\link=2)$ is obtained. 

From here, we consider the arm's telescopic extension in the orthogonal direction (Fig.~\ref{fig:combined_primitives}c--top/right, \add{SI Video 2}). First, we note that as the arm crosses the spiral plane, its equivalent knot passes through itself twice, causing writhe and link to decrease by four\cite{Fuller:1978} (discontinuity in Fig.~\ref{fig:combined_primitives}d), with the handedness (sign) of $\twist$ switching from positive to negative. 
Second, as long as the LM (the strongest group) remain fully contracted, we expect the arm tip to approximately maintain its tangent, thus approaching the boundary conditions of Fig.~\ref{fig:combined_primitives}b. 
This implies a constant $\link \approx 2$ (SI), which we numerically confirm (Fig.~\ref{fig:combined_primitives}d) and exploit to force twist to increase at the expense of writhe (Eq.~\ref{eq:CFW}) by contracting R-OM, thus unfolding the helix (Fig.~\ref{fig:combined_primitives}c).

\noindent\textbf{Automatic control of sucker alignment.} Our topological interpretation also allows us to understand how the positioning of the sucker line, a key functionality \cite{Grasso:2008}, is automatically determined by the arm architecture. Indeed, since suckers and longitudinal muscles run parallel to the arm axis at a radial offset, they are physical auxiliary curves. Thus, their common axial rotation is determined by the oblique muscles. Then, as the arm morphs into a 3D shape upon LM contractions, the extend to which suckers are exposed outward (convex side) or tucked inward (concave side, Fig.~\ref{fig:combined_primitives}a-2) is governed by which set of longitudinal muscles is activated. As illustrated in Fig.~\ref{fig:combined_primitives}c (top row), LM on the suckers’ side cause the suckers to always face inward, a useful feature during grasping. It is instead sufficient to contract the opposite set of LM to expose all suckers outward (for sensing), while retaining the same arm morphology (Fig.~\ref{fig:combined_primitives}c, bottom row).

This mechanism is critically enabled by the co-activation of both oblique and longitudinal muscles. Indeed, LM alone can form any 3D axial shape by controlling writhe, but they cannot (fully) determine the orientation of the auxiliary curves, which requires additional control over twist (OM).  In the SI we show how local contractions by LM only can form helical shapes, albeit at the cost of increasing problem complexity and forgoing control of the suckers. Thus, the presence of oblique muscles within the arm architecture allows to both simplify control and command orientation, potentially justifying their evolutionary emergence. We conclude by emphasizing that while the exact muscle activations  the octopus employs remain unknown, our model compactly captures a range of observations \add{(SI Video 6)}, directly yielding insights into robotic design and control.

\begin{figure}[t!]
    \centering
    \includegraphics[width=\linewidth]{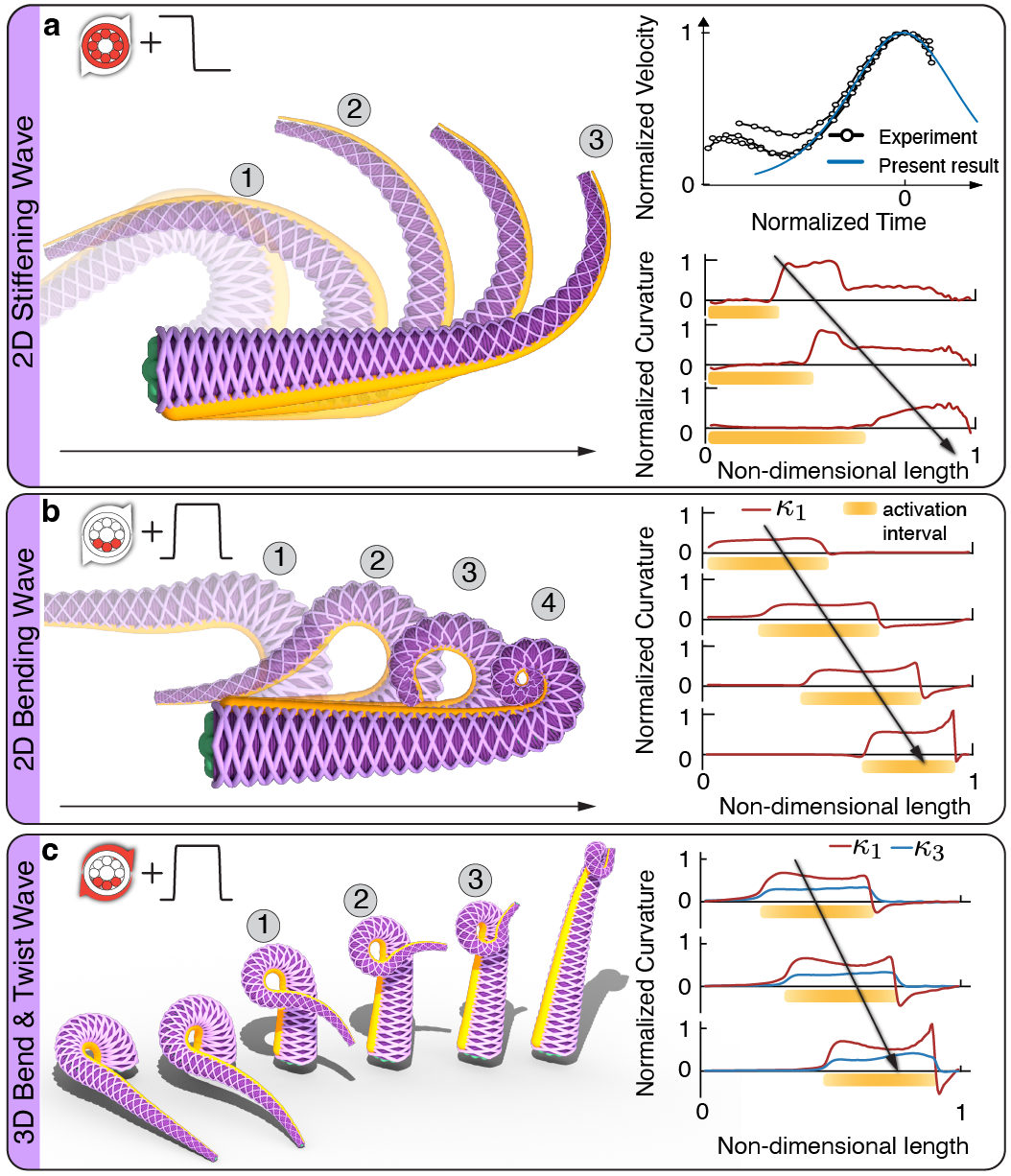}
    \caption{
    \textbf{Active geometric transport.} \textbf{(a)} A stiffening wavefront (\wavesymbol) generates a reaching motion by propagating a bend ($\kappa_1$) along the arm (activation details in SI). 
    Simulations capture both peak bend velocity (30 cm/s) and normalized bend velocity profile (SI) as experimentally determined by Gutfreund et al. \cite{Gutfreund:1996,Gutfreund:1998}. 
    \textbf{(b)} A localized bend primitive is injected into the arm at its base and transported by a pulse (\pulsesymbol) of LM muscle activation (activation details in SI). The bending ($\kappa_1$) profile shows the curvature packet transport in relation with LM activations. 
    \textbf{(c)} Generalization of kernel activation and transport to 3D. Example of pulse (\pulsesymbol) of LM (bend) and OM (twist) activations (details in SI), and resulting  traveling of curvature packets ($\kappa_1,\kappa_3$). All motions are available in \add{SI Video 3}.
    }
    \label{fig:inject_transport}
    \vspace{-20pt}
\end{figure}

\noindent\textbf{Geometric transport via waves of muscle actuation.} 
Next, we augment uniform actuation strategies with the transport of localized activations along the arm. We are inspired by planar bending propagation observed in octopuses during reaching and fetching behaviors\cite{Gutfreund:1996, Gutfreund:1998, Sumbre:2001, Sumbre:2005, Sumbre:2006, Richter:2015}. This stereotyped motion, presumed to simplify neuromuscular control, consists of a localized bend that travels along the arm before exiting at the tip. Because of its prevalence in many octopus activities, this behavior has been particularly well characterized, through EMG recordings\cite{Gutfreund:1998, Sumbre:2001}, kinematic data\cite{Richter:2015}, and dynamic modeling
\cite{Yekutieli:2005-1,
Hanassy:2015}. 
These studies have converged on the hypothesis that the bend is formed and transported by a stiffening wavefront of isometric LM and TM co-contractions. 
By incorporating a similar traveling wavefront (\wavesymbol) of muscle activation, we recapitulate planar reaching motions and experimentally observed bend propagation velocities\cite{Gutfreund:1996,Gutfreund:1998} (Fig.~\ref{fig:inject_transport}a, SI), further validating our model while suggesting potential control generalizations.

\begin{figure*}[t!]
    \vspace{-3pt}
    \centering
    \includegraphics[width=\textwidth]{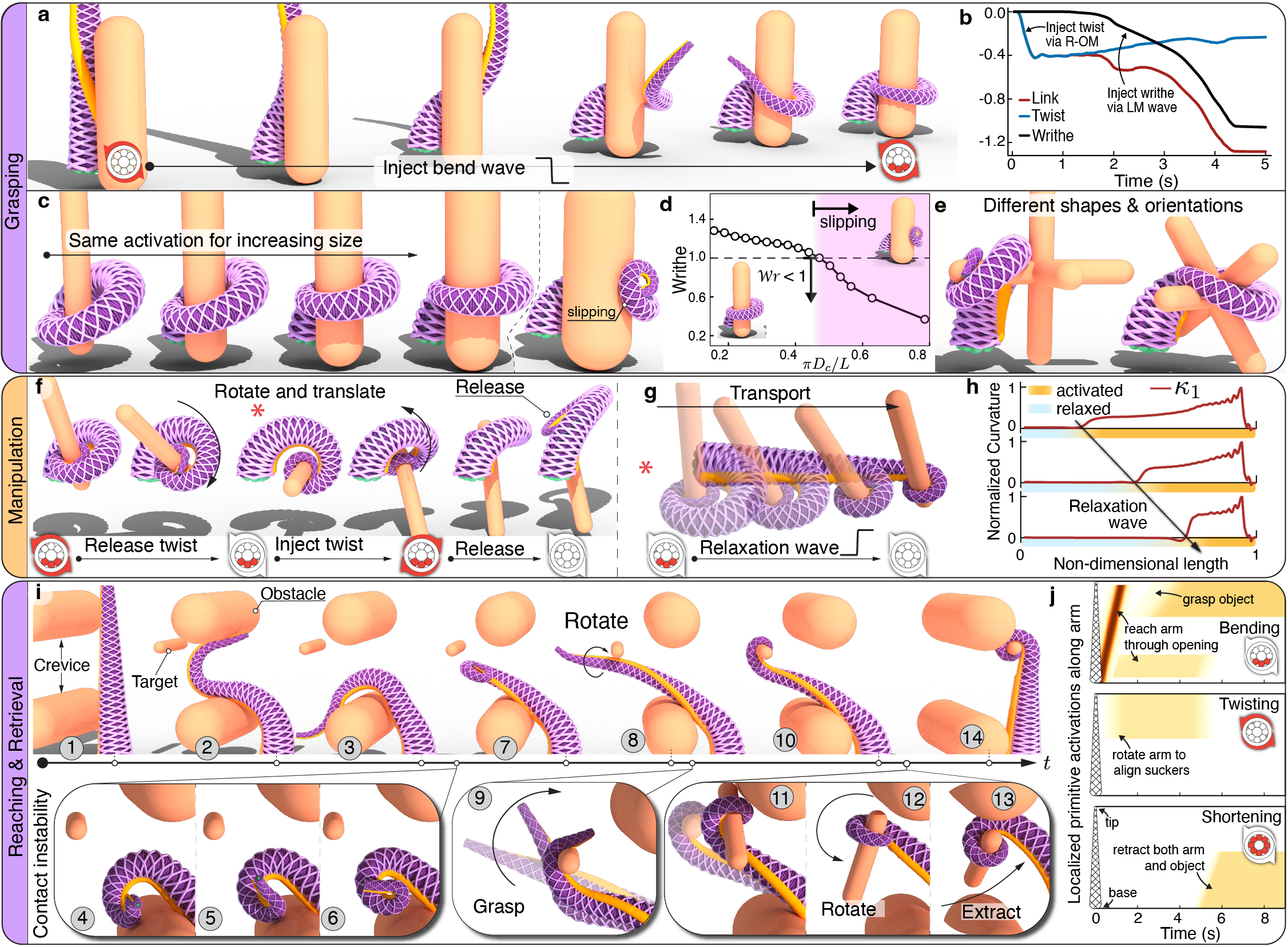}
    \caption{
    \textbf{Complex behaviors: object grasping, manipulation, and retrieval.} \textbf{(a)} Grasping an object by injecting twist into the base and coiling via a bending wave. 
    \textbf{(b)} Evolution of link, writhe, and twist of the arm during grasping motion. 
    \textbf{(c)} The same muscle activation pattern successfully grasps cylinders of increasing diameter \add{(SI Video 4)}. 
    \textbf{(d)} Writhe in an arm of length $L$ grasping objects of diameter $D_c$. When $|\writhe| \lesssim 1$ the arm begins to slip (pink region). 
    \textbf{(e)} The same muscle activation of (a) also allows the arm to grasp objects of different shapes and orientations. 
    \textbf{(f)} A grasped object is manipulated by sequentially releasing R-OM and activating L-OM at the base, to transport the object into the opposite plane before being released (muscle relaxation) \add{(SI Video 5)}. 
    \textbf{(g)} The arm from (f), in the configuration denoted by the red star, transports the obstacle away from the base via an LM wave of relaxation \add{(SI Video 5)}. 
    \textbf{(h)} As the relaxation wave travels towards the tip, the arm’s passive elasticity straightens the arm, and bending curvature $(\kappa_1)$ becomes localized  tightening grip. 
    \textbf{(i)} Arm reaching through a crevice to grasp and retrieve a target object. Insets: (steps 4-6) arm’s compliance accommodates imperfect reaching and solid obstacles by buckling out of plane; (step 7) grasping of the target; (steps 11-13) retrieval of the object where compliance corrects for imprecise control \add{(SI Video 6)}. 
    \textbf{(j)} Spatiotemporal activation maps for (i).
    }
    \label{fig:reaching_grasping}
    \vspace{-14pt}
\end{figure*}

Indeed, moving beyond the octopus to issues of robotic actuation and control, the stiffening wave mechanism can be extended to generic muscle activation kernels. These are defined as spatially-compact sets of muscle activations resulting in localized reconfigurations of the arm. Within this framework, we can revisit planar bend propagation as a traveling pulse (\pulsesymbol) of one-sided LM contractions, to inject local curvature. Figure~\ref{fig:inject_transport}b illustrates how this strategy allows the formation and transport of a tightening spiral which, as we will see later, can be used for object displacement. The approach can be readily generalized to 3D local structures, enhancing manipulation and reconfiguration range. In Fig.~\ref{fig:inject_transport}c we inject a pulse (\pulsesymbol) of bending and twist at the base of the arm, via the co-contraction of LM and OM. As the pulse propagates along the arm, a localized corkscrew structure is seen to form, translate, and tighten. 

\noindent\textbf{Grasping, manipulation, and interfacial interactions.} 
The injection and transport of muscle activation kernels, together with uniform actuation templates, provide a framework to understand grasping and manipulation in soft arms (natural or artificial).

We first consider the grasping of a cylindrical post, as illustrated in Fig.~\ref{fig:reaching_grasping}a. To firmly grasp the object, the arm must form one or more coils around it. Geometrically, this implies $\writhe > 1$, however, writhe alone is insufficient to ensure grasping, as the coil must be correctly oriented to encompass the object. We achieve this by revisiting the helix formation process of Fig.~\ref{fig:combined_primitives}. By injecting a localized twist (via R-OM contraction) in the first half of the arm from the base, we turn the suckers towards the target, before contracting the longitudinal muscles on the suckers’ side using a wavefront (\wavesymbol) of activation (Fig.~\ref{fig:reaching_grasping}b). This causes the arm to dynamically wrap around the object while appropriately aligning the suckers inwards. Here, the use of a muscle contraction wave is found to be significantly more robust than a uniform activation strategy. Indeed, in the first case, the distal portion of the arm progressively winds around the post without intersecting it, while in the second case, the arm tends to coil too early causing the distal end to make contact, preventing wrapping (SI). To demonstrate the reliability of the chosen approach, we test the ability of the exact same muscle actuation sequence to deal with objects of different size (Fig.~\ref{fig:reaching_grasping}c), shape, and orientation (Fig.~\ref{fig:reaching_grasping}e). In the absence of a solid interface, the activation sequence of Fig.~\ref{fig:reaching_grasping}a,b causes the arm to tightly coil up. It is the obstacle’s presence that passively informs and modulates the arm’s morphology, as it complies and conforms to the presented target. This simplifies and robustifies control, leading to successful grasping across scenarios. Confirming our initial geometric intuition, as the obstacle’s circumference approaches the length of the arm, writhe decreases and, for $|\writhe| \lesssim 1$, we begin to observe grasping failure, with the arm’s distal end slipping off the obstacle (Fig.~\ref{fig:reaching_grasping}d). While octopuses supplement grasping with their suckers to securely attach to the substrate \cite{Grasso:2008}, here we do not explicitly include this effect. However, we do consider friction (both static and kinetic) and in the SI show how it affects the onset of slipping, with low friction causing grasping failure. Thus, interfacial force modulation via suckers (or engineered analogs \cite{Sholl:2019,Xie:2020}) is an avenue to expand operational range.

Once grasped, an object can be manipulated by activating muscles at locations not in contact. For example, activating muscles at the base of the arm allows the arm to rotate or translate the object while maintaining its grasp. In Fig.~\ref{fig:reaching_grasping}f, the arm rotates the object into a horizontal position by first relaxing its R-OM (thus bringing the arm into a planar spiral, similar to Fig.~\ref{fig:combined_primitives}), and then transports it across the spiral plane by contracting its L-OM. To release its grasp, the arm can simply relax, letting passive elasticity loosen its grip. Passive elasticity can also facilitate object transport. In Fig.~\ref{fig:reaching_grasping}g, an LM relaxation wavefront travels out from the base of the arm, transporting the object along. As the longitudinal muscles relax, the arm uncurls  while the initially injected curvature becomes increasingly localized in the distal section (Fig.~\ref{fig:reaching_grasping}h), enabling the arm to maintain its grasp.

Octopuses regularly probe and reach through small crevices to retrieve objects of interest\cite{Richter:2015, Richter:2016}. We conclude by combining the lessons learned so far to enable a similar behavior in our model (Fig.~\ref{fig:reaching_grasping}i,j). Motion is initiated by injecting a bending wave from the base, as in Fig.~\ref{fig:inject_transport}b, although this time the first portion of the longitudinal muscles are left contracted to maintain the arm positioned towards the opening (steps \circled{1}-\circled{3}). As the arm attempts to reach through, it encounters the obstacles that define the opening, however, thanks to its compliance, this disturbance is naturally accommodated, with the distal end of the arm deforming out of plane to slip past the solid boundary (\circled{4}-\circled{6}). Note that the suckers have been so far exposed outwards (LM active on same side), as experimentally observed in octopuses \cite{Richter:2015,Wang:2022}. As the arm makes its way across the crevice, twist is injected to align the arm’s suckers with the target (\circled{7}) for grasping (\circled{8}, \circled{9}), which is enacted by bending the distal end via localized LM contractions (on the suckers’ side). Once grasped, the injected twist is released, rotating both arm and object by 180$^\circ$ (\circled{10}-\circled{12}), before all longitudinal muscles of the proximal and medial regions are activated (shortening primitive). This pulls the arm back out through the opening to successfully extract the target despite obstacle collisions (\circled{13}, \circled{14}),  which are again passively dealt with via arm compliance. Notably, this entire motion sequence is accomplished by combining only three actuation primitives (bending wave, twisting, and shortening), with spatiotemporal profiles reported in Fig.~\ref{fig:reaching_grasping}j.


\vspace{4pt}
\noindent\textbf{Conclusion.} By combining medical imaging and biomechanical data with rod-based modeling of heterogeneous, fibrous structures, the uniquely complex architecture of the octopus arm, and the mechanical program it embodies, is decoded through the lens of topology and geometry. A control framework for grasping and manipulation is revealed, replicating prototypical behaviors experimentally observed. Gleaned insights not only advance our understanding of muscular hydrostats and provide testable hypotheses, but also inform translatable principles for the design and control of soft robots that seek to match the dexterity and finesse of their natural counterparts. Finally, this work is significant in terms of modeling from imaging, whereby automation and application to musculoskeletal structures promises opportunities in patient-specific medical and assistive care.

\printbibliography[segment=0]
\newrefsegment 

\end{document}